\documentclass[conference]{IEEEtran}
\ifCLASSINFOpdf
\else
\fi
\def\BibTeX{{\rm B\kern-.05em{\sc i\kern-.025em b}\kern-.08emT\kern-.1667em\lower.7ex\hbox{E}\kern-.125emX}}

\usepackage{bbm}
\usepackage{nicefrac}
\usepackage{siunitx}
\usepackage{array,framed}
\usepackage{booktabs}
\usepackage{
  color,
  float,
  epsfig,
  wrapfig,
  graphics,
  graphicx,
}
\usepackage{textcomp,amssymb}
\usepackage{setspace}
\usepackage{latexsym,fancyhdr,url}
\usepackage{enumerate}
\usepackage[linesnumbered,ruled]{algorithm2e}
\usepackage{algpseudocode}
\usepackage{graphics}
\usepackage{subfigure}
\usepackage{xparse} 
\usepackage{xspace}
\usepackage{multirow}
\usepackage{csvsimple}
\usepackage{balance}

\usepackage{amsmath}

\usepackage{
  tikz,
  pgfplots,
  pgfplotstable
}
\usepackage{hyperref}

\usetikzlibrary{
  shapes.geometric,
  arrows,
  external,
  pgfplots.groupplots,
  matrix
}

\pgfplotsset{compat=1.9}

\usepackage{mathtools,}

\DeclareMathAlphabet{\mathcal}{OMS}{cmsy}{m}{n}

\DeclareGraphicsExtensions{%
    .png,.PNG,%
    .pdf,.PDF,%
    .jpg,.mps,.jpeg,.jbig2,.jb2,.JPG,.JPEG,.JBIG2,.JB2}

\usepackage{xparse}
\newcommand{\bnm}{\begin{newmath}}
\newcommand{\enm}{\end{newmath}}

\newcommand{\bea}{\begin{eqnarray*}}%
\newcommand{\eea}{\end{eqnarray*}}%

\newcommand{\bne}{\begin{newequation}}
\newcommand{\ene}{\end{newequation}}

\newcommand{\bal}{\begin{newalign}}
\newcommand{\eal}{\end{newalign}}

\newenvironment{newalign}{\begin{align}%
\setlength{\abovedisplayskip}{4pt}%
\setlength{\belowdisplayskip}{4pt}%
\setlength{\abovedisplayshortskip}{6pt}%
\setlength{\belowdisplayshortskip}{6pt} }{\end{align}}

\newenvironment{newmath}{\begin{displaymath}%
\setlength{\abovedisplayskip}{4pt}%
\setlength{\belowdisplayskip}{4pt}%
\setlength{\abovedisplayshortskip}{6pt}%
\setlength{\belowdisplayshortskip}{6pt} }{\end{displaymath}}

\newenvironment{newequation}{\begin{equation}%
\setlength{\abovedisplayskip}{4pt}%
\setlength{\belowdisplayskip}{4pt}%
\setlength{\abovedisplayshortskip}{6pt}%
\setlength{\belowdisplayshortskip}{6pt} }{\end{equation}}

\newcounter{ctr}

\newcounter{mytable}
\def\mytable{\begin{centering}\refstepcounter{mytable}}
\def\endmytable{\end{centering}}

\newcounter{myfig}
\def\myfig{\begin{centering}\refstepcounter{myfig}}
\def\endmyfig{\end{centering}}

\newlength{\saveparindent}
\newlength{\saveparskip}
\newcommand{\E}{{\rm I\kern-.3em E}}

\newcommand{\secref}[1]{\mbox{Section~\ref{#1}}}

\renewcommand{\eqref}[1]{\mbox{Equation~(\ref{#1})}}

\def \part {part}

\renewcommand{\paragraph}[1]{\vspace*{6pt}\noindent\textbf{#1}\;}

\def \blackslug{\hbox{\hskip 1pt \vrule width 4pt height 8pt
    depth 1.5pt \hskip 1pt}}
\def \qed{\quad\blackslug\lower 8.5pt\null\par}

\newcounter{mynote}[section]

\newcommand\ignore[1]{}

\newcounter{rcnote}[section]

\newcounter{mrnote}[section]

\newcounter{fknote}[section]

\newcounter{anote}[section]

\DeclareMathSymbol{\mlq}{\mathord}{operators}{``}
\DeclareMathSymbol{\mrq}{\mathord}{operators}{`'}

\newcommand{\rhf}[2]{R_{f, \gamma}}

\DeclareDocumentCommand{\edist}{o o}{
  \ensuremath{
    \IfNoValueTF{#1}{{d}}{{\sf d}(#1,#2)}
  }
}

\newcommand{\olrk}[1]{\ifx\nursymbol#1\else\!\!\mskip4.5mu plus 0.5mu\left(\mskip0.5mu plus0.5mu #1\mskip1.5mu plus0.5mu \right)\fi}

\NewDocumentCommand{\indseq}{ O{1} O{r} }{{#1}\ldots {#2}}

\begin{document}
%
\title{CLMIA: Membership Inference Attacks via Unsupervised Contrastive Learning}

\author{\IEEEauthorblockN{Depeng Chen}
\IEEEauthorblockA{ Anhui University\\
}
\and
\IEEEauthorblockN{Xiao Liu}
\IEEEauthorblockA{Anhui University}
\and
\IEEEauthorblockN{Jie Cui}
\IEEEauthorblockA{Anhui University}
\and
\IEEEauthorblockN{Hong Zhong}
\IEEEauthorblockA{Anhui University}
}


%


\IEEEoverridecommandlockouts
\makeatletter\def\@IEEEpubidpullup{6.5\baselineskip}\makeatother
\IEEEpubid{\parbox{\columnwidth}{
This paper is an extended version of our work published at ACM CCS 2023, "Poster: Membership Inference Attacks via Contrastive Learning". Jie Cui is the corresponding author of this paper.
}
\hspace{\columnsep}\makebox[\columnwidth]{}}

\maketitle

\begin{abstract}
 Since machine learning model is often trained on a limited data set, the model is trained multiple times on the same data sample, which causes the model to memorize most of the training set data. Membership Inference Attacks (MIAs) exploit this feature to determine whether a data sample is used for training a machine learning model. However, in realistic scenarios, it is difficult for the adversary to obtain enough qualified samples that mark accurate identity information,  especially since most samples are non-members in real world applications. To address this limitation, in this paper, we propose a new attack method called CLMIA, which uses unsupervised contrastive learning to train an attack model without using extra membership status information. Meanwhile, in CLMIA, we require only a small amount of data with known membership status to fine-tune the attack model. Experimental results demonstrate that CLMIA performs better than existing attack methods for different datasets and model structures, especially with data with less marked identity information. In addition, we experimentally find that the attack performs differently for different proportions of labeled identity information for member and non-member data. More analysis proves that our attack method performs better with less labeled identity information, which applies to more realistic scenarios.

\end{abstract}

\maketitle

\section{Introduction}
\label{sec:intro}

In recent years, the development of artificial 
intelligence (AI) has been in full swing. Machine 
learning (ML) is the core part of AI, and it has a wide range of applications in many fields, such as image recognition, natural language processing, and 
recommendation systems. With the continuous 
development of ML technologies, especially deep learning, massive amounts of data can be utilized more fully and efficiently. This is because t raining ML models require a large amount of user data, which often involves personal privacy content, such as financial information\cite{chen2022predicting}, medical data information\cite{zhang2022membership}, location information\cite{pyrgelis2017knock} and other 
personally sensitive data.
\par
\begin{figure}[htbp]
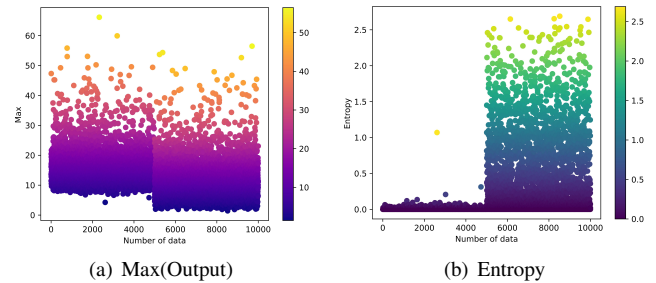
 
    \centering
    \subfigure[Max(Output)]{
        \label{Fig.sub.1}
        \includegraphics[scale=0.018]{images/MAxss_bar-eps-converted-to.pdf}
    }
    \subfigure[Entropy]{
        \label{Fig.sub.2}
        \includegraphics[scale=0.018]{images/ENxss_bar-eps-converted-to.pdf}
    }
    \caption{\textbf{(a) Maximum output value of the last layer of the target model. The first 5000 samples are members, and the last 5000 are non-members. (b) Information entropy of the posterior probabilities of the target model.}}
    \label{fig.1}
\end{figure}

Prior studies\cite{carlini2019secret,shokri2017membership,long2018understanding,irolla2019demystifying,salem2019ml} have been shown that ML models largely remember the data in 
the training set, leading to the disclosure of 
private information in the training set. A membership inference attack (MIA) is an attack that determines whether data samples are used to train an ML model, and this attack leads to the disclosure of private information about the ML model.
\par

In the last few years, to test the leakage of membership information of the ML models, most existing MIAs\cite{shokri2017membership,hu2022membership,liu2022membership,sablayrolles2019white} have focused on improving attack methods through more fine-grained analysis or reducing the background knowledge and computing power required to conduct an attack. Specifically, most of MIA methods train attack models to perform inference attacks based on the posterior of the target model, which has significantly different outputs for members and 
non-member data, mainly because ML models are 
trained on a limited dataset, so the models are trained 
multiple times on the same data sample. And this feature gives the model 
sufficient ability to remember the data samples in the training set, resulting in the model performing differently against the training set data and the non-training set data. Therefore, for these different performances, MIA can detect the difference in performance and infer member and non-member data, resulting in the disclosure of private information on the training dataset. As shown in Figure \ref{fig.1}, We can easily see that the target model has significantly different output results for member and non-member data, which leads the adversary to effectively distinguish whether the data samples are in the training set of the target model. However, this approach tends to misclassify non-member data as member data, leading to a high false-positive rate (FPR). Carlini \textit{et al.} \cite{carlini2022membership} pointed out that MIA should reduce the occurrence of high FPR and follow the true-positive rate (TPR) at the Low FPR strategy. 
Besides, this distinction between members and non-members requires the data identified to be labeled in advance to train the attack model text, which is difficult in a realistic scenario where a large amount of data with general membership information is available\cite{hui2021practical}. Therefore, in this paper, we propose an attack method to train the attack model using unsupervised contrastive learning, which requires less data with labeled membership information and reduces the adversary's a priori knowledge.
\par
This paper proposes a novel membership inference attack against ML models called \textbf{Contrastive Learning MIA--CLMIA}. It takes advantage of contrastive learning as unsupervised learning that does not need to focus on the tedious details of instances, learns to distinguish between data only on the feature space, and is more generalizable. To mount CLMIA, the adversary has the target dataset to infer information about members and non-members in the dataset and a small amount of information about the data samples known to the attacker. First, the adversary obtains the dataset needed to train the attack model. Since it is a contrastive learning strategy, positive and negative samples are required to train an attack model. However, the target model only has one input for one data sample into one output, so to solve this problem, we train a shadow model. The shadow model and the target model have the same model structure, and the only difference is that the shadow model has more dropout layers than the target model \cite{gao2021simcse}. So by adding a dropout layer to the shadow model, a pair of positive samples is obtained. For obtaining negative samples, we use all the samples in the same training batch as negative samples of that data sample by referring to \cite{chen2020simple}. The attacker then uses a supervised method to fine-tune the attack model via a small sample with known data identification information. To ensure that the parameters of the unsupervised training attack model are not changed, all parameters of the attack model are frozen here. And only the parameters of the last added MLP layer can be changed during the supervised training process. In the end, the training of the attack model is completed, and the inference of the membership information is only made through the posterior of the target model as input to the attack model so that the attack model can infer the membership information of the data samples. In general, this paper makes the following contributions:

\begin{itemize}
    \item To our best knowledge, we are the first to use the method of unsupervised contrastive learning as a means of membership inference attack, called CLMIA.
    
    \item We demonstrate that CLMIA can be applied to more realistic scenarios and achieve better attack performance, especially in the case of insufficient data of labeled identity information.
    
    \item We apply CLMIA to various datasets and model structures and conduct ablation experiments to show the efficient performance of our attack model.
\end{itemize}

\textbf{Roadmap.} In~\secref{sec:relwork}, we introduce the background knowledge of membership inference attacks and contrastive learning.  \secref{sec:methodology} presents the threat model, design intuition, and details of the CLMIA attack. We demonstrate the performance of our scheme through a large number of experiments and evaluate the strategy in~\secref{sec:eval} and explore the critical factors affecting the attack performance through ablation experiments in~\secref{sec:abl}. Finally, we discuss related work in~\secref{sec:related} and conclude the paper in~\secref{sec:conclusion}.



\section{Background}
\label{sec:relwork}

\subsection{Membership Inference Attack}
Membership inference attacks lead to the disclosure of data sample privacy in the training datasets by determining whether the data samples are used for training machine learning models. Specifically, MIA detects whether the data sample $x_i$ comes from the training dataset $D_{train}=\left\{(x_1,y_1),(x_2,y_2),...,(x_n,y_n)\right\}$ of the target model ${f_\vartheta} $ by high-precision prediction, thereby threatening the privacy of data providers. The following function defines a typical MIA:
\begin{equation}
     \mathcal{A}:{K,f_ \vartheta,x} \to \left\{0 \ or \ 1 \right\}
\end{equation}

Where $\mathcal{A}$ is the attacker, $K$ is the degrees of the target model the attacker possesses, i.e., the attacker's prior knowledge. $x$ is the data sample waiting for inference, and $f_\vartheta$ is the target model. If the output is 0, it is judged that the sample $x$ does not belong to the training data set $D_{train}$,i.e.$x \notin D_{train}$, otherwise $x \in D_{train}$.

\paragraph{Shadow model based MIA.}
Shadow model-based MIA was first proposed by \cite{shokri2017membership}, which is implemented by training multiple shadow models with the same structure and training data set as the target model. This scheme imitates the behavior of the target model through multiple shadow models to achieve the privacy of the target model training set without direct access to the target model. Still, the attack performance of this method relies on the transferability of the shadow model and the target model and requires training multiple shadow models, which increases the attack overhead. To reduce the problem of high attack overhead due to introducing multiple shadow models, \cite{salem2019ml} proposes MIA based on a single shadow model, where attacker $\mathcal{A}$ only needs to train one shadow model, which not only saves computational attack overhead but also achieves the same attack performance as multiple shadow models.
\par
In addition, \cite{long2018understanding, irolla2019demystifying, 8728167, sablayrolles2019white, nasr2019comprehensive, song2019privacy, li2022user} conducted a more fine-grained analysis of it, as well as being applied in different scenarios, such as industrial IoT \cite{chen2020practical}, recommender systems \cite{zhang2021membership}, semantic segmentation \cite{he2020segmentations,zhang2022label} and speech models 
\cite{shah2021evaluating,miao2021audio} etc.

\paragraph{Threshold-based MIA.}
In threshold-based comparison MIA, most of the proposed methods do not use shadow models, but obtain the output of the target model directly, followed by statistical analysis of it, and predict the membership relationship by comparing it with the threshold value set in advance. Li and Zhang \cite{li2021membership} propose an attack schema that by comparing the cross-entropy loss $CE_{Loss}$, the method considers the loss in training set smaller than in the test set. Therefore when the $CE_{Loss}$ is smaller than the threshold $\tau$ set in advance, it is judged as a member. Otherwise, it is a non-member. Furthermore, Rahimian \textit{et. al} \cite{rahimian2021differential} present a sampling attack to generate perturbed samples of data samples $x$ by a perturbation function $pert()$, which argues that at a specific level of perturbation, member data do not easily change their labels compared to non-member data because training data are further away from the decision boundary than non-training data samples. Besides, \cite{del2022leveraging} pointed out to generate adversarial samples by adding perturbed noise to change their original prediction labels and, finally, to determine the membership relationship by measuring the size of the added noise. However, this one-size-fits-all decision approach only applies to simple decision boundaries, and no effective attack can be implemented for target models with complex decision boundaries. Therefore,  it is essential to propose a more fine-grained attack to evaluate each data sample's impact on the dataset's distance by differential comparison \cite{hui2021practical}. Specifically, it determines the membership of a sample in the target dataset by moving the sample in the target dataset and observing the change in the difference between the target dataset and the non-member dataset.

\subsection{Contrastive Learning}
\label{sec:overview}
As an unsupervised learning approach, contrastive learning can learn standard features between instances of the same type and distinguish differences between non-similar cases. It only needs to learn to distinguish data on the feature space at the abstract semantic level, making the model and its optimization more straightforward and generalizable. \cite{chen2020simple} proposed a typical contrastive learning framework for data sample $x$ to generate two positive samples $x_i$, $x_j$ by data augmentation, and the rest in the same batch are used as negative samples. Subsequently, $x_i$ and $x_j$ go through a base encoder network $f(.)$, and a projection head $g(.)$ are trained to maximize agreement using a contrastive loss, respectively. After the training, $g(.)$ is discarded, and only $f(.)$ is used to predict downstream tasks. The contrastive loss function is defined as

\begin{equation}
    \ell_{i, j}=-\log \frac{\exp \left(\operatorname{sim}\left(\boldsymbol{z}_{i}, \boldsymbol{z}_{j}\right) / \tau\right)}{\sum_{k=1}^{2 N} \mathbbm{1}_{[k \neq i]} \exp \left(\operatorname{sim}\left(\boldsymbol{z}_{i}, \boldsymbol{z}_{k}\right) / \tau\right)}
    \label{equal 2}
\end{equation}

Where $z_i$ and $z_j$ are the outputs of $x_i$ and $x_j$ respectively after a projection head $g(.)$, $sim(.)$ is the cosine similarity, $\tau$ a temperature parameter, $\mathbbm{1}_{[k \neq i]} \in \left\{0 \ or \ 1 \right\}$ is an indicator function evaluating to 1 iff $k \neq i$.

In addition, \cite{Chen_2021_CVPR} proposed contrastive learning frameworks using Siamese networks and without negative samples. Specifically, their architecture takes as input two randomly augmented views $x_1$ and $x_2$ from an image $x$. The two views are processed by an encoder network $f(.)$ and a projection MLP head, denoted as $h(.)$, which achieves the prediction of $z_{2} \triangleq f\left(x_{2}\right)$ by $p_{1} \triangleq h\left(f\left(x_{1}\right)\right)$.

In this paper, we use a contrastive learning model to extract features of members and non-members on an unlabeled target dataset.

\section{Attack Methodology}
\label{sec:methodology}

In this section, we introduce the methodology of CLMIA. We first define the threat model based on the attacker's knowledge of the target model and then introduce the design intuition of our attack method. And the details of our attack method CLMIA are proposed at last.

\subsection{Threat Model}
In this paper, we set the attacker's knowledge as a black box, which does not know the structure of the target model and only has access to the posterior probabilities of the target model. The attacker only has an unlabeled target dataset $D_t$, which infers the membership of its dataset and a few labeled datasets $D_l$ that have identified members and non-members. This is realistic because the attacker can get a target dataset for reasoning about the membership in the dataset and has a small amount of data with annotated membership information.

\subsection{Design Intuition}
Existing MIAs use a supervised approach to train the attack model and require a large dataset with identified member states for training the attack model, which essentially limits the generalization ability of the attack model. In realistic scenarios, it is difficult to obtain such a large number of datasets with labeled members and non-members. To mitigate this obstacle, we propose a contrastive learning framework-- CLMIA, which has good generalization capability through contrastive learning and does not require labeled data during the training phase of the attack model and only uses a small amount of labeled data when fine-tuning the attack model.

Our attack method distinguishes between members and non-members only by the posterior probability of the target model output. For the output posterior probability of a data sample, two similar positive samples are generated using data augmentation. Then the model is trained in an unsupervised manner by using a loss function for contrast learning. By contrast learning, we expect to make member data closer to each other and different types of data further away from each other. Thus, the attack model can distinguish between the member and non-member data efficiently.

\begin{figure*} [htbp]       
	\centering
    \includegraphics[scale=0.64]{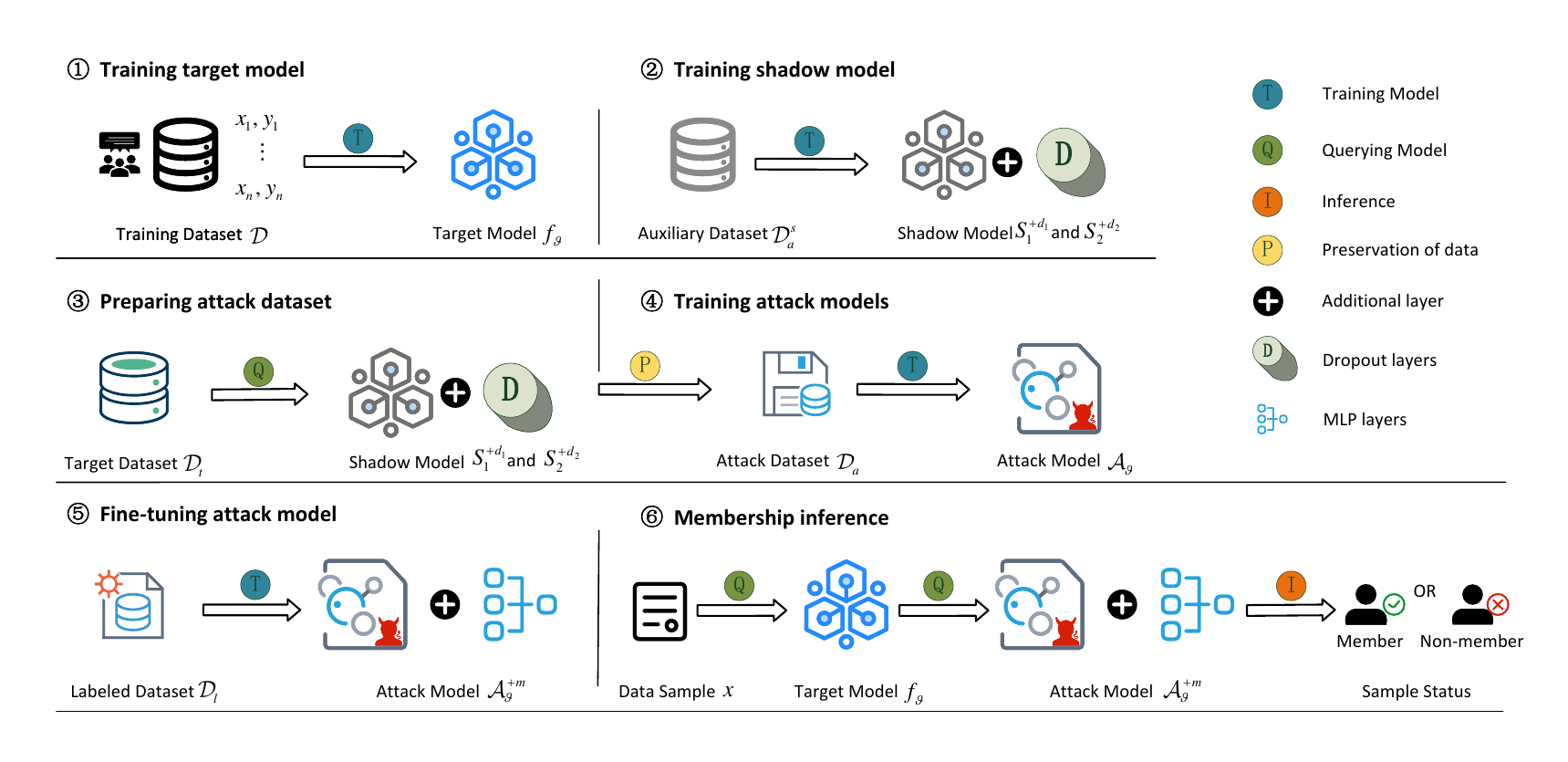}  
	\caption{System Model of Our CLMIA}   
	\label{FIG:Figure 1}
\end{figure*}

\subsection{Attack Method}
Inspired by contrastive learning not requiring labeled data and the trained model having better generalization ability, we propose a new membership inference attack called contrastive learning MIA, CLMIA. To the best of our knowledge, this is the first time unsupervised learning has been applied to train an attack model. In CLMIA, the attacker uses a black-box approach to access the target model. The attacker does not need to know the structure and internal parameters of the target model but only the output of the last layer of the target model. This is the same way as the current mainstream MIA approach ~\cite{10.1145/3548606.3560684} to access the target model. Besides, the attacker has a target dataset $D_t$ for which membership information has not been inferred and a limited number of datasets $D_l$ for which membership information has been annotated. This is all the information the adversary has about the target model, which is easier to obtain in a real scenario.

In addition, the positive and negative samples are needed to train the model using contrastive learning. To solve this issue, We adopt the dropout layer. Specifically, the adversary first adds the dropout layer to the output of the last layer of the target model as the positive samples $x_i$ and $x_j$ generated from the data sample $x$. By doing this, the adversary can obtain positive samples and train the attack model using contrastive learning. After obtaining positive samples, the adversary can fine-tune the attack model obtained from the above training based on a small number of labeled datasets $D_l$. Finally, the attack model can infer the target dataset's membership state only by the output of the last layer of the target model.

\par

\begin{algorithm}[t]
    \SetAlgoLined
    
    \caption{Unsupervised training of attack models}\label{algorithm.1}
    \KwIn{$f_\vartheta,\mathcal{D}_t, \tau, \varepsilon, S_1^{+d_1}, S_2^{+d_2}$}
    \KwOut{$\mathcal{A}_\vartheta$}
    \textbf{Preparing attack training dataset}\;
    \For{$x \in \,\mathcal{D}_t$}{
        $S_1^{+d_1} (x), S_2^{+d_2}(x) \rightarrow x_i, x_j \rightarrow p$\;
        $p^* \leftarrow p \oplus max(p) \oplus entropy(p)$\;
        
        $\mathcal{D}_a \leftarrow p^* + \mathcal{D}_a$\;
    }
    \textbf{Training attack model}\;
    \For{$x_i, x_j \in \, \mathcal{D}_a$}{
        $z_i, z_j \leftarrow \mathcal{A}_\vartheta(x_i,x_j)$\;
        $sim(z_i, z_j)={z_i^T \cdot z_j} / {\|z_i\|\|z_j\|}$\;
        $\mathcal{L}_{i, j}=-\log \frac{\exp \left(\operatorname{sim}\left(z_{i}, z_{j}\right) / \tau\right)}{\sum_{k=1}^{2 N} \mathbbm{1}_{[k \neq i]} \exp \left(\operatorname{sim}\left(z_{i}, z_{k}\right) / \tau\right)}$\;
        
        $\vartheta_{i+1}=\vartheta_{i}-\varepsilon \sum_{x \in B} \nabla_{\vartheta} \mathcal{L}_{i,j}$\;
        
    }
    \textbf{return} $\mathcal{A}_\vartheta$

\end{algorithm}

To summarize, the overall pipeline of CLMIA is illustrated in Figure \ref{FIG:Figure 1}. It consists of five stages: training the target model, training the shadow model, training the attack model, fine-tuning the attack model, and membership inference. The detailed implementation of each specific phase is as follows:

\paragraph{Training the Target Model.}
First, several users form a data provider to share their data and combine them into a training dataset $D$, which is provided to the target model $f_\vartheta$ as training data. Figure \ref{FIG:Figure 1} part 1 shows the training target model phase.

\paragraph{Training the Shadow model.}
In the second step, by adding dropout layers to the target model and using different dropout rates $d_1$ and $d_2$, we obtain two shadow models $S_1^{+d_1}$ and $S_2^{+d_2}$. It is shown in the following Equation:
\begin{equation}
    S_1 + d_1 \rightarrow S_1^{+d_1} ; 
    S_2 + d_2 \rightarrow S_2^{+d_2}
    \label{equal 4}
\end{equation}
where $S_1$ and $S_2$ have the same structure as the target model $f_\vartheta$, $d_1$ and $d_2$ denote the addition of dropout layers with different dropout rates.

Thus, we can obtain two positive samples $x_i$ and $x_j$ corresponding to the samples $x$ according to the shadow models to prepare the dataset for training the attack model.

\paragraph{Training the Attack Model.}
The training attack model phase includes two main parts: preparing the attack training dataset and the training attack model phase. The details are shown in Algorithm \ref{algorithm.1}.
\begin{itemize}
    \item \paragraph{Preparing the attack training dataset.}
In the training attack model phase, the attacker obtains the target dataset $D_t$, which contains membership and non-membership data. The attacker wants to infer membership information in this dataset. First, the attacker based on the shadow models $S_1^{+d_1}$ and $S_2^{+d_2}$ obtained in the previous phase. Then, each $x$ data in the target dataset $D_t$ is sequentially passed through the $S_1^{+d_1}$ and $S_2^{+d_2}$ to obtain the respective positive samples $x_i$ and $x_j$, where $x_i$ and $x_j$ are expressed as a vector of posteriors, denoted as $p=[p_1,...,p_C]$, and $C$ is the number of classes. Meanwhile, the attacker also computes the posterior probability $p$ by performing a statistical calculation and finally obtains $p^*$ stored in the dataset $D_a$ as the training data for the attack model.
\begin{equation}
    p^*=p \oplus max(p) \oplus entropy(p)
    \label{equal 3}
\end{equation}
where $max(p)$ denotes taking the maximum value of $p$ and $entropy(p)$ is the information entropy of $p$.

    \item \paragraph{Training attack models.}
Since the training data $D_a$ for the attack model has been obtained above, the attacker then trains the attack model using a comparative learning strategy. The goal of contrastive learning is to learn an encoder that encodes similar data of the same kind and makes the encoding results of different data classes as different as possible. Thus, it can extract information that distinguishes the membership status without the attacker knowing the data membership information in advance. The training attack model uses to NT-Xent(the normalized temperature-scaled cross-entropy loss)\cite{chen2020simple}, as in equation \ref{equal 2}.
\end{itemize}

\paragraph{Fine-tuning the Attack Model.}
The above-trained completed attack model already has the ability to extract member and non-member features. However, the adversary is not yet able to obtain explicit information about the membership of the sample. To this end, the adversary adds a layer of MLP to the attack model on a small number of labeled datasets $D_l$. The MLP layer is trained by supervised learning only while keeping the parameters of the attack model unchanged. By fine-tuning the attack model, the final attack model can infer and output membership information. The details are shown in Algorithm \ref{algorithm.2}.

\begin{algorithm}[t]
    \SetAlgoLined
    
    \caption{Supervised fine-tuning of attack models}\label{algorithm.2}
    \KwIn{$\mathcal{A}_\vartheta$, $\mathcal{D}_l$, $\varepsilon$ }
    \KwOut{$\mathcal{A}_\vartheta^{+m}$}
    \textbf{Add MLP layer}\;
    $\mathcal{A}_{\vartheta}.parameters  \leftarrow False $
    \tcp{freeze all layers of $\mathcal{A}_{\vartheta}$}\
    $\mathcal{A}_\vartheta^{+m} \leftarrow\mathcal{A}_\vartheta + MLP$\;
    \textbf{Training MLP layers only}\;
    \For{$x, target \in \mathcal{D}_l$}{
        $status \in (0 \, or \, 1) \leftarrow A_\vartheta^{+m}(x)$\;
        $\vartheta_{i+1}=\vartheta_{i}-\varepsilon \sum \nabla_{\vartheta} \mathcal{L}\left(status, target\right)$\;
        
    }
    \textbf{return}  $A_{\vartheta}^{+m}$

\end{algorithm}


\paragraph{Membership Inference.}
Finally, the adversary can infer the identity information of each sample in dataset $D_t$ only from the posterior output $p^*$ of the target model $f_\vartheta$, and the adversary can perform successful inference even for samples that are not in dataset $D_t$. The main reason is that the attack model trained using contrastive learning has better generalization ability.

\section{Evaluation}
\label{sec:eval}

\subsection{Experimental Setup}
In this section, we implement extensive experiments using multiple model structures and datasets to evaluate our attack method CLMIA and compare it with other representative schemes.

\paragraph{Datasets.}
we conduct our experiments on the following three public image datasets:

\begin{itemize}
    \item \textbf{CIFAR-10:}
The dataset includes 50,000 training data and 10,000 test data, and the images are 3-channel 32*32 color RGB images divided into 10 classes, including aircraft, birds, cats, frogs, and other classes.
    \item \textbf{CIFAR-100:}
The dataset consists of 50,000 training data and 10,000 test data. The images are 3-channel 32*32 color RGB images divided into 100 classes, and these 100 classes are divided into 20 superclasses, each containing 5 more fine-grained labels.
    \item \textbf{STL-10:}
The dataset consists of 5000 training data and 8000 test data with 3-channel 96*96 color RGB images, divided into 10 classes, each with fewer labeled training examples than CIFAR-10.

\end{itemize}

\paragraph{Models.}
For the choice of model structure, we exploit the following three models to test the performance of CLMIA, including simple CNN, VGG-19, and Resnet-18 for the target model. The performance of different model structures on different data sets, i.e., the model's training accuracy and testing accuracy, is expressed as the degree of overfitting of the model. As shown in Table \ref{tab:model performance}.

\begin{table}[]
\caption{\textbf{Performance of the target model on different data sets}}
    \centering
    \begin{tabular}{c|ccc}
    \toprule
     Dataset& Model Structure& Train acc& Test acc\\ \midrule
     \multirow{3}*{CIFAR-10}& CNN& 100\%& 77.6\%\\
                            & VGG-19& 100\%& 92.7\%\\
                            & Resnet-18& 100\%& 88.7\%\\ \midrule
    \multirow{3}*{CIFAR-100}& CNN& 99.9\%& 43.7\%\\
                            & VGG-19& 99.9\%& 71.5\%\\
                            & Resnet-18& 99.9\%& 67.9\%\\ \midrule
    \multirow{3}*{STL-10}& CNN& 100\%& 58.4\%\\
                        & VGG-19& 86.2\%& 56.9\%\\
                        & Resnet-18& 100\%& 77.5\%\\ 
     
\bottomrule
\end{tabular}
    \label{tab:model performance}
\end{table}

\paragraph{Metrics.}
We evaluate the effectiveness of our approach through the following metrics:
\begin{itemize}
    \item \textbf{Balanced Accuracy.}
    This metric means that there are an equal number of member and non-member data samples in the test dataset. This gives an accuracy rate of 50 \% if the attacker is guessing at random.
    
    \item \textbf{$F_1$ Score.}
    It is used to measure the performance of the classification model, taking into account both the precision and recall of the classification model.
    \item \textbf{ROC.}
    Receiver Operating Characteristic (ROC) curve is exploited to compare the ratio of true-positives to false-positives to evaluate the classifier's performance. 
\end{itemize}
\begin{table*}
	\centering
	\caption{\textbf{Attack performance of different attacks against ResNet-18 trained on three datasets. Additional results for the other two model architectures with a similar pattern can be found in Section \ref{A1}.}}
	\begin{tabular}{c|c|ccccccccc}
		\toprule
		\multicolumn{2}{c}{\multirow{2}[4]{*}{\textbf{Attack method}}} & \multicolumn{3}{c}{\textbf{Balanced accuracy}} & \multicolumn{3}{c}{\textbf{$F_1$ Score}} \\
		\cmidrule(r){3-5} \cmidrule(r){6-8} \multicolumn{2}{c}{} & \multicolumn{1}{c}{CIFAR-10} & \multicolumn{1}{c}{CIFAR-100} & \multicolumn{1}{c}{STL-10} & \multicolumn{1}{c}{CIFAR-10} & \multicolumn{1}{c}{CIFAR-100} & \multicolumn{1}{c}{STL-10} \\
		
        \midrule
  	\multirow{4}[2]{*}{Based on thresholds} & Top1  &0.501  &0.521 & 0.502     & 0.667     &0.676      & 0.667      \\
		& Prediction entropy  & 0.581     & 0.801     & 0.631     &0.705 &0.834 & \textbf{0.730 }    \\
		& Modified entropy & 0.568     & 0.701     & 0.624     & 0.698     &0.770      & 0.727      \\
		& Predicted loss  &0.542      & 0.641     & 0.583     &0.686      & 0.736     &0.706    \\
        \cmidrule(r){1-2}
        \multirow{1}[1]{*}{Based on labels} & Prediction correctness &0.560   &0.660   &0.610      &0.694    & 0.746     &0.719      \\
        \cmidrule(r){1-2}
		\multirow{1}[1]{*}{Based on the attack model} & NN attack  &0.659  & \textbf{0.970} &0.628 & 0.768     &\textbf{0.970 }     &  0.640     \\
         \cmidrule(r){1-2}
		\textbf{Ours} & \textbf{CLMIA} & \textbf{0.668} &0.894  & \textbf{0.637}     & \textbf{0.779}   & 0.903     & 0.691   \\
		\bottomrule
	\end{tabular}%
	\label{tab:res_c100}
\end{table*}%

\begin{figure*}
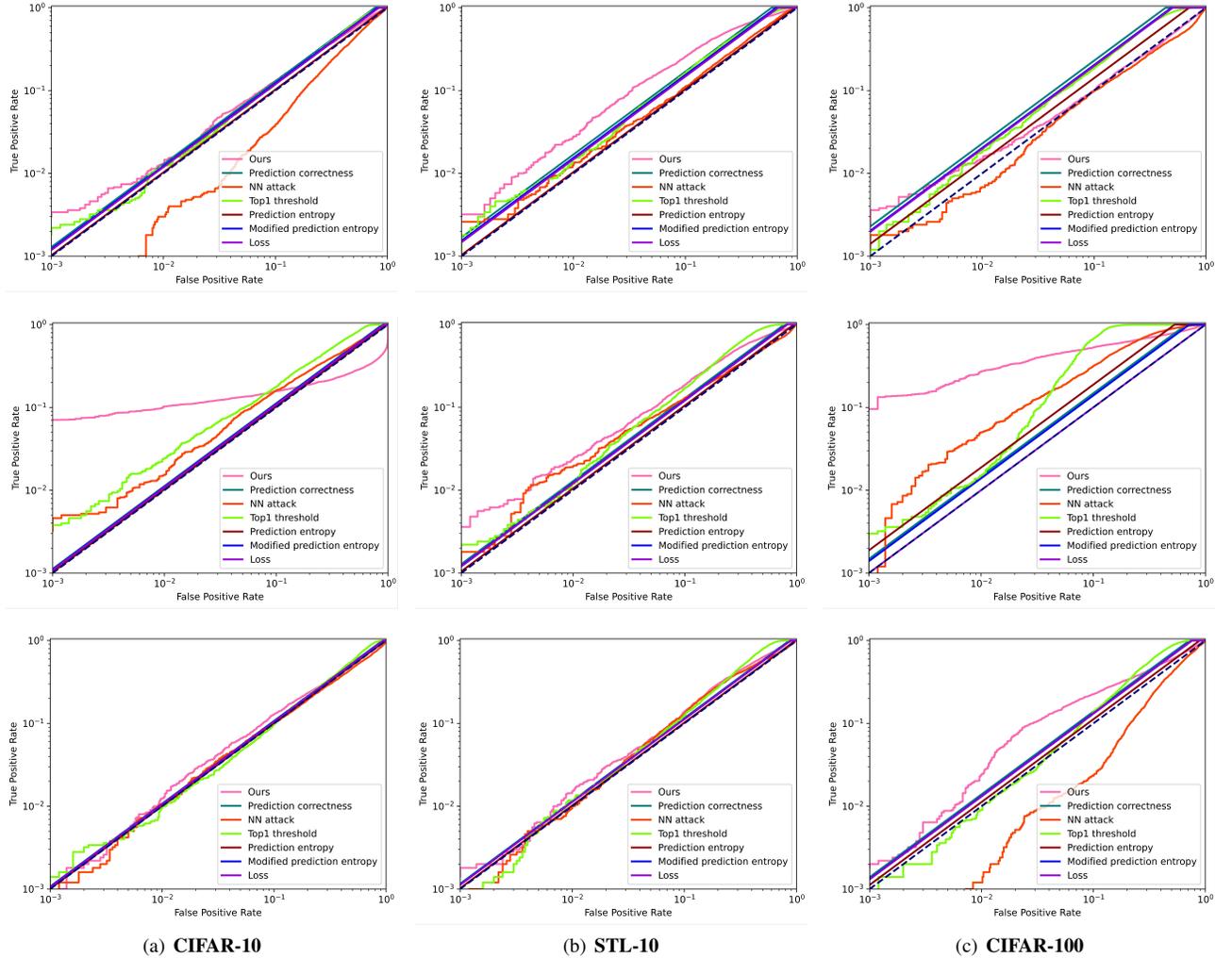

	\centering
        \subfigure{
        \includegraphics[scale=0.03]{images/c10_cnn_s.pdf}
        }\subfigure{
        \includegraphics[scale=0.03]{images/stl10_cnn_s.pdf}
        }\subfigure{
        \includegraphics[scale=0.03]{images/c100_cnn_s.pdf}
        }   
        \subfigure{
        \includegraphics[scale=0.03]{images/c10_res_s.pdf}
        }\subfigure{
        \includegraphics[scale=0.03]{images/stl10_res_s.pdf}
        }\subfigure{
        \label{FIG.sub.3a}
	\includegraphics[scale=0.03]{images/c100_res18_s.pdf}
        }    
        \setcounter{subfigure}{0}
        \subfigure[\textbf{CIFAR-10}]{
        \label{FIG.sub.3b}
	\includegraphics[scale=0.03]{images/c10_vgg_s.pdf}
        }\subfigure[\textbf{STL-10}]{
        \label{FIG.sub.3c}
	\includegraphics[scale=0.03]{images/stl10_vgg_s.pdf}
        }\subfigure[\textbf{CIFAR-100}]{
        \label{FIG.sub.3d}
	\includegraphics[scale=0.03]{images/c100_vgg_s.pdf}
        }
	\caption{\textbf{ROC curves for attacks on three different datasets and three model architectures (from top to bottom: Simple CNN, Resnet-18, VGG-19)}}
	\label{fig.fpt_res_c100}
\end{figure*}

\begin{figure*} [ht]       
	\centering
        \subfigure{
        \includegraphics[scale=0.262]{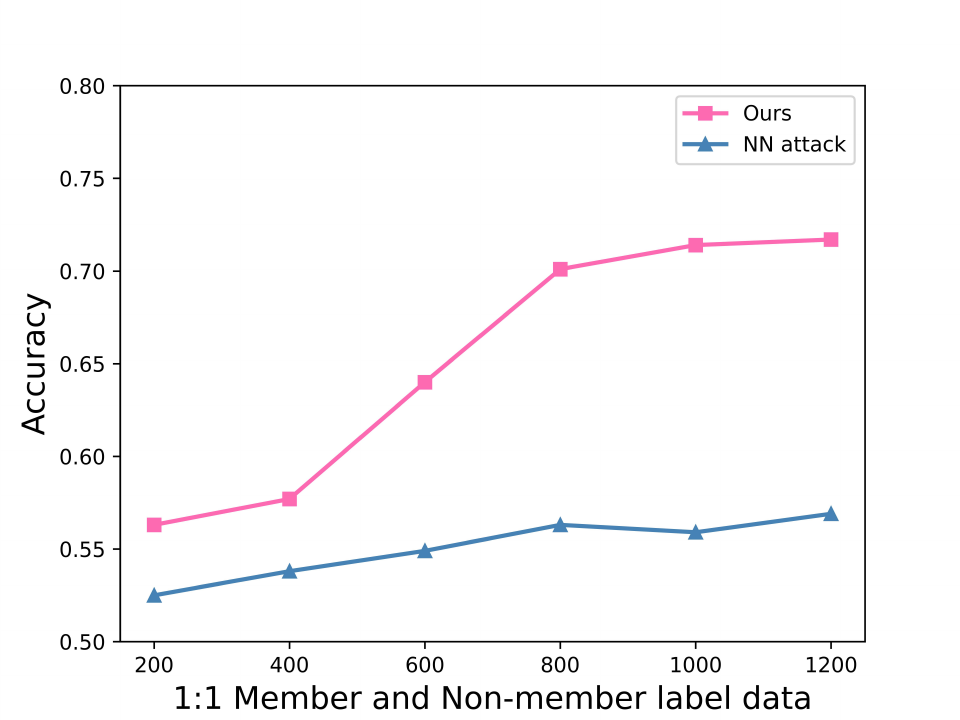}
        }\subfigure{
        \includegraphics[scale=0.262]{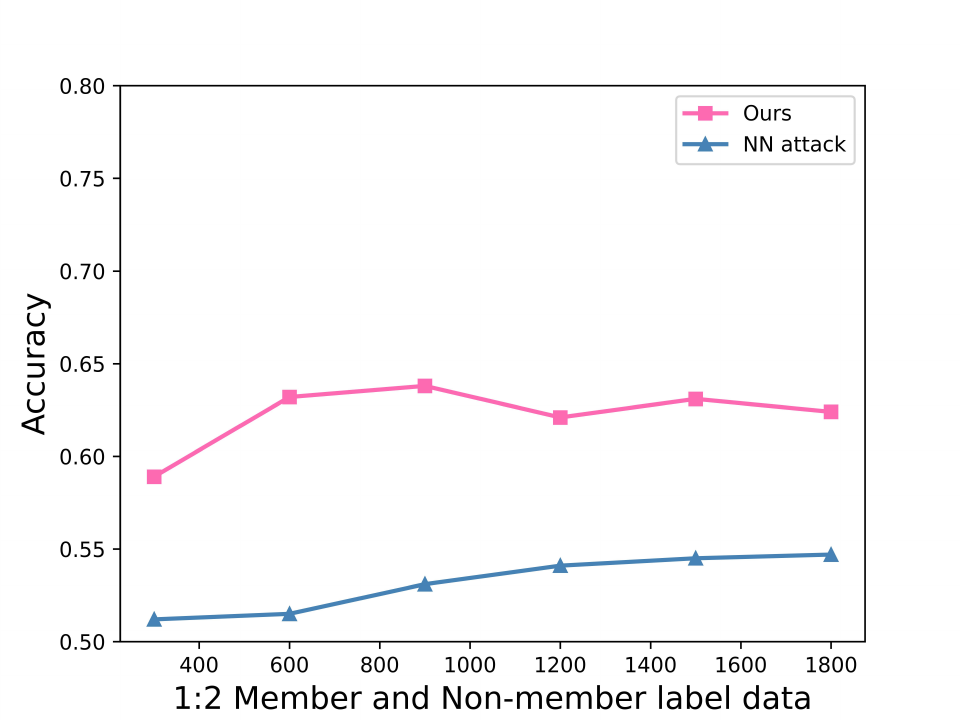}
        }\subfigure{
	\includegraphics[scale=0.262]{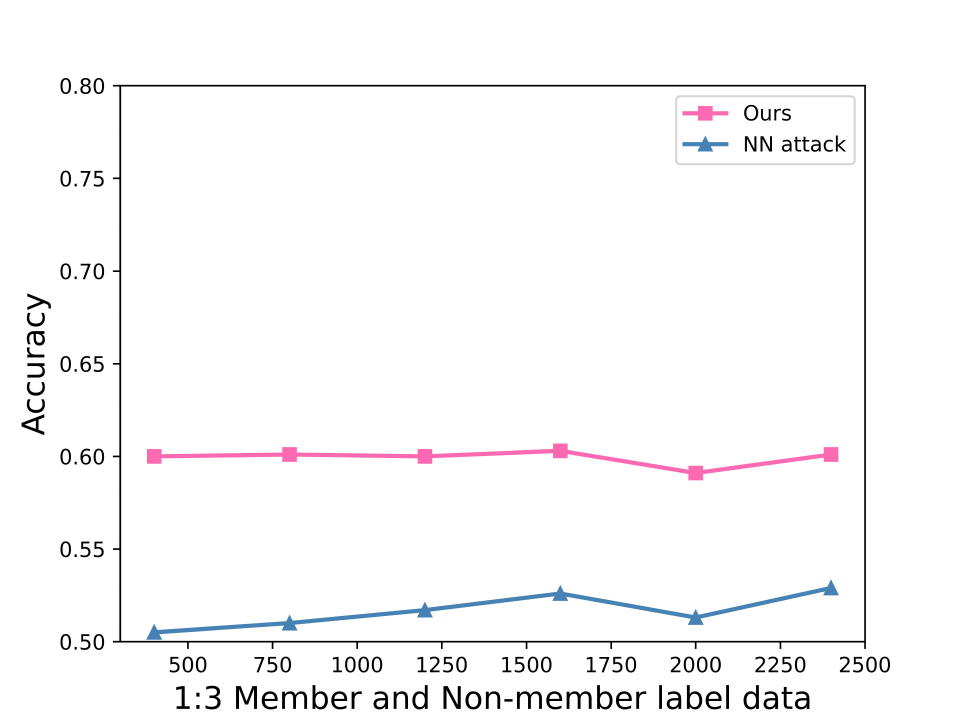}
	}
        \subfigure{
	\includegraphics[scale=0.262]{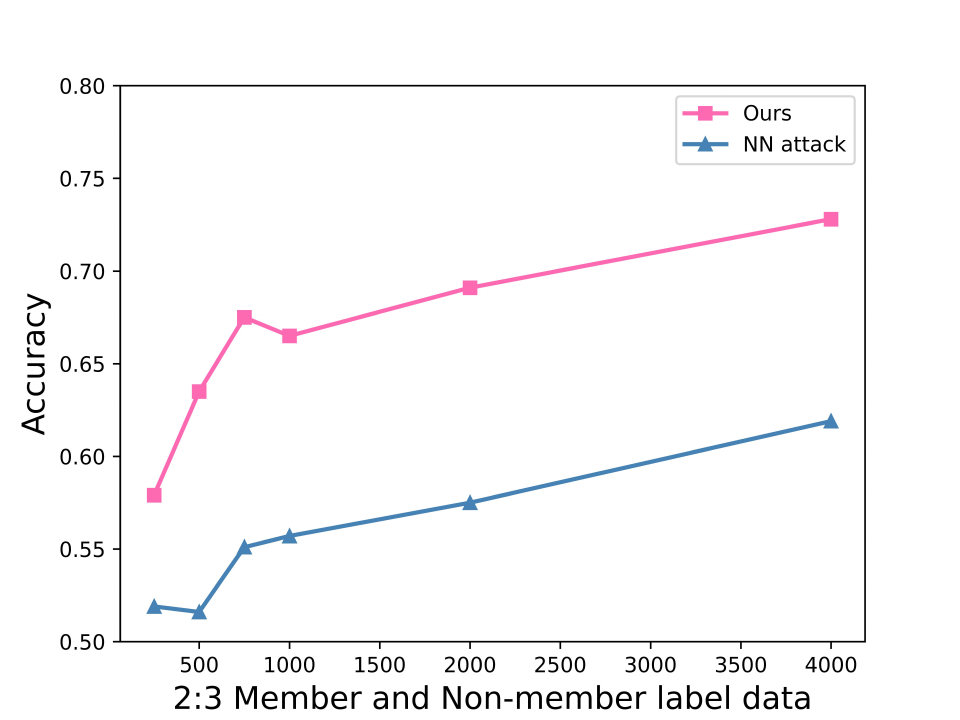}
         }
	\caption{\textbf{The effect of the size of the labeled dataset on the balanced accuracy in the CIFAR-100 dataset, with a model structure of Resnet-18.\\}}
	\label{fig.labled set size}
\end{figure*}

\paragraph{Attack Baselines.}
In this part, we show the effectiveness of our CLMIA by comparing with five existing representative MIA methods \cite{leino2020stolen,salem2019ml,song2021systematic}, where the attack methods are as follows:
\begin{itemize}
    \item \textbf{Prediction correctness.}
     Lein \textit{et al.} \cite{leino2020stolen} point out making the sample a member when the model outputs the correct predictive label and a non-member otherwise. This attack evaluates existing models against MIAs well but performs poorly against attacks on models with powerful generalization capabilities.
    
    \item \textbf{NN attack.} Salem \textit{et al.} \cite{salem2019ml} propose to train the attack model by using a single shadow model combined with the posterior of the target model.
    
    \item \textbf{Top1 threshold.}
    Salem \textit{et al.} \cite{salem2019ml} calculate the maximum value of the target model by its output posterior and judge it as a member if the value is greater than the pre-set threshold of $\tau$. Otherwise, it is a non-member.
    
    \item \textbf{Prediction entropy.}
    In addition to Top1 threshold method, Salem \textit{et al.} \cite{salem2019ml} also propose to determine member status based on the posterior of the target model by calculating the information entropy of each data sample and comparing it with a threshold value set in advance.
    
    \item \textbf{Modified prediction entropy.}
     Song \textit{et al.} \cite{song2021systematic} argue that MIA based on predicted entropy may lead to the misclassification of members and non-members and therefore improves on predictive entropy.

    \item \textbf{Predicted losses.}
    Yeom \textit{et al.} \cite{yeom2018privacy} state that the samples in the training set should have a smaller prediction loss than the non-training set and therefore proposes to compare the MIA by the prediction loss values.
    
\end{itemize}

\subsection{Experimental Results}

In this section, we show the attack performance of CLMIA, comparing it with 6 different attack approaches in the context of three datasets and three other model structures. The performance of the target model is shown in Table \ref{tab:model performance}.

We evaluate different attacks in a black box scenario and classify the attack methods. The experimental results are shown in Figure \ref{fig.fpt_res_c100}, where our attack methods can achieve the best performance in most scenarios with the least knowledge the adversary possesses on the low-FPR regime. In addition, as shown in Table \ref{tab:res_c100}, our attack method can perform better than other attack methods in the evaluation approach of balanced accuracy and $F_1$ score.

\section{Ablation Study}
\label{sec:abl}
In this section, we analyze several essential factors that affect the performance of the attack. First, we pass through only the fully Connected layers (FC layers) as a comparison to demonstrate that learning, by contrast, enables the target model to learn the distinction between members and non-members. We then explore the performance of the attack when the adversary knows only a limited amount of membership information. Finally, we also analyze the effect of dropout rates and different temperature parameters in the loss function on the attack performance.

\subsection{Only FC Layer}
\begin{table}[ht]
\caption{\textbf{Attack performance of CLMIA and Only FC layer with balance accuracy on different target model structures and datasets}}
    \centering
    \begin{tabular}{c|ccc}
    \toprule
     Dataset& Model Structure& CLMIA& Only FC layer \\ \midrule
     \multirow{3}*{CIFAR-10}& CNN& \textbf{65.2\%}& 56.5\%\\
                            & VGG-19& \textbf{61.9\%}& 51.3\%\\
                            & Resnet-18& \textbf{66.8\%}& 54.7\%\\ \midrule
    \multirow{3}*{CIFAR-100}& CNN& \textbf{73.8\%}& 63.4\%\\
                            & VGG-19& \textbf{67.9\%}& 62.4\%\\
                            & Resnet-18&\textbf{89.4\%}& 80.1\%\\ \midrule
    \multirow{3}*{STL-10}& CNN& \textbf{65.5\%}& 60.5\%\\
                        & VGG-19& \textbf{63.7\%}& 57.4\%\\
                        & Resnet-18& \textbf{64.9\%}& 54.1\%\\ 
     
\bottomrule
\end{tabular}
    \label{tab:clmia_fc_tab}
\end{table}

\begin{figure} [ht]
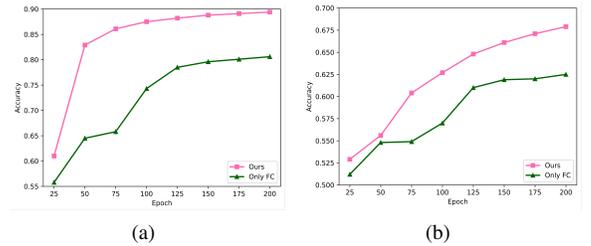
       
	\centering
	\subfigure[]{
	\label{Fig.sub.fcres}
	\includegraphics[scale=0.015]{images/res_fc_ours.pdf}
	}\subfigure[]{
	\label{FIG.sub.fcvgg}
	\includegraphics[scale=0.015]{images/vgg_fc_ours.pdf}
	}
	\caption{\textbf{The performance of CLMIA and Only FC layer on the test dataset during training. Where (a) denotes performance on dataset CIFAR-100 and model structure Resnet-18. (b) denotes performance on dataset CIFAR-100 and model structure VGG-19.}}
	\label{clmia_fc}
\end{figure}

In this part, we demonstrate that the attack model trained by the contrastive learning approach can distinguish between member and non-member data rather than relying on the FC layer during the fine-tuning phase of the attack model. To this end, we only use the FC layer with the same model and dataset, compare the CLMIA attack method with the attack model through experiments, and also use the same training strategy and hyperparameters.

The results are summarized in Figure \ref{clmia_fc} and Table \ref{tab:clmia_fc_tab}. As expected, the adversary can perform effective attacks using CLMIA methods instead of relying only on the FC layer.

\subsection{Labeled Member Or Nonmember Set Size}

\begin{figure} [ht]
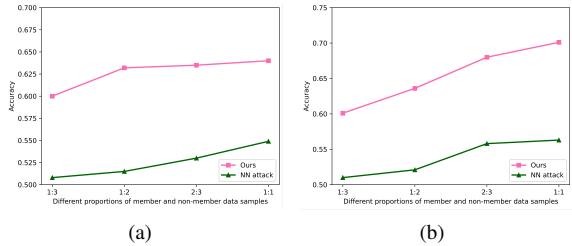
       
	\centering
	\subfigure[]{
	\label{Fig.sub.600size}
	\includegraphics[scale=0.015]{images/600size.pdf}
	}\subfigure[]{
	\label{FIG.sub.800size}
	\includegraphics[scale=0.015]{images/800size.pdf}
	}
	\caption{\textbf{Impact of different member and non-member data ratios on attack balance accuracy for the same label data size, where (a) has a label status data size of 600 and (b) is 800.}}
	\label{fig.size_por}
\end{figure}

A necessary condition for membership inference attacks is that the adversary can train an attack model from a few known membership status data and that the attack model can infer the membership information in the whole training dataset. In a realistic scenario, the adversary does not have sufficient data to satisfy the quality requirement for membership information. Therefore, we explore the attack performance of CLMIA and NN attacks with a small amount of labeled membership data and different proportions of members and non-members.

The experimental results are shown in Figure \ref{fig.labled set size}. We conduct experiments using the number of labeled membership data and non-membership data at 1:1, 1:2, 1:3, and 2:3, respectively. We find that CLMIA performs better than NN attack regarding balance accuracy when using the same number and ratio of labeled membership information, even when there is little information. This may be attributed to the fact that CLMIA uses contrast learning to train the attack model while making the attack model have better generalization ability. In particular, we also discovered that the performance of attack balance accuracy varies when different proportions of members and non-members are labeled with identity information during the training of the attack model. Figure \ref{fig.size_por} shows a more explicit representation. We can easily find that even for attack models trained with the same amount of labeled status data, the different proportions of different member and non-member data impact the attack balance accuracy. It is easy to find that the higher the proportion of membership status data to the overall amount of data, the higher the performance of its balancing accuracy. In other words, the attack performance increases with the increase in the membership proportion.

\subsection{Dropout Rates}

\begin{figure} [ht]      
	\centering
        \subfigure{
        \includegraphics[scale=0.024]{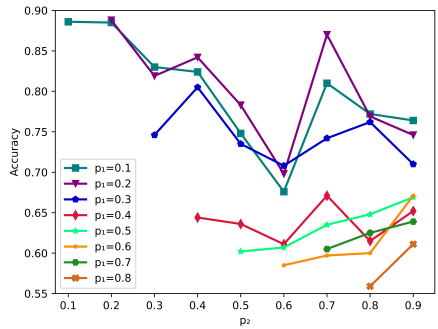}
        }
	\caption{\textbf{The impact of different dropout rates of the shadow model on the attack balance accuracy in CIFAR-100 dataset and model structure of Resnet-18, where $p_1$ denotes the dropout used by shadow model $S_1^{+d_1}$ and $p_2$ is the dropout rates used by shadow model $S_2^{+d_2}$.}}
	\label{fig.dropout}
\end{figure}

In contrastive learning, selecting positive samples is crucial to the excellence of model training. If an unsuitable positive sample is selected, it will affect the accuracy and generalization ability of model training or even lead to the collapse of the training model. Therefore, it is necessary for CLMIA to choose the optimal dropout rates to obtain the appropriate positive samples for training the attack model.

We can see from Figure \ref{fig.dropout} that in most cases, a higher dropout rate generates positive samples for training the attack model corresponds to the lower attack performance in balance accuracy. However, there are some exceptions, where most models show a significant performance improvement when $p_2$ equals 0.7. In addition, there is a significant improvement for $p_1$ greater than 0.4 and $p_2$ at 0.9. Overall, in the CIFAR-100 dataset and the target model with model Resnet18, the adversary's training for shadow models $s_1$ and $s_2$ can use dropout rates of 0.1 to obtain positive samples and train the resulting attack models with superior performance. This reason may be that the attack model is made to achieve the distinction between members and non-members as much as possible while ensuring that the member and non-member features are not lost.

\subsection{Temperature parameters}

\begin{figure} [ht]       
	\centering
        \subfigure{
        \includegraphics[scale=0.024]{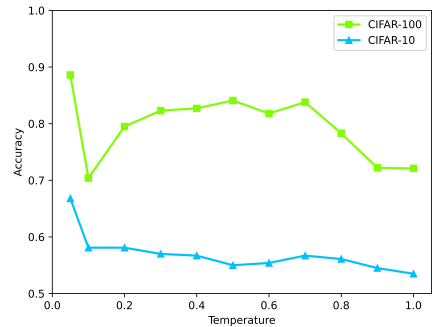}
        }
	\caption{\textbf{Effect of different temperature coefficients $\tau$ in  the Resnet-18 model structure}}
	\label{fig.temperature}
\end{figure}

The loss function, one of the most fundamental and critical elements of machine learning, is used to express the degree of disparity between predictions and actual data. In contrastive learning, the loss function is used to measure the distance between positive and negative samples. In the CLMIA loss function, the temperature parameter $\tau$ is used to measure the sensitivity of the loss to negative samples. The smaller the $\tau$, the more the loss will tend to give a more significant penalty to negative samples with more significant similarity. The larger the $\tau$, the greater the loss tends to give more minor penalties to negative samples with more remarkable similarity. And $\tau$ needs to be chosen according to different task scenarios. Therefore, we obtain the optimal CLMIA attack performance by comparing other temperature parameters $\tau$.

As can be seen from Figure \ref{fig.temperature}, as expected, a significant temperature parameter makes the loss function give a minor penalty for negative samples, resulting in an attack model that does not effectively distinguish between member and non-member data, and the optimal temperature parameter t in CLMIA has a value of 0.05 in the Resnet-18 model structure.

\section{Related Work}
\label{sec:related}

\subsection{Membership Inference Attacks}

Membership inference attack (MIA) is a new type of attack that leads to the disclosure of user privacy during machine learning and has attracted extensive academic attention in recent years. Shokri \textit{et al.} \cite{shokri2017membership} first propose deep learning classification tasks for MIAs, which use multiple shadow models to simulate the behavior of the target model. And the output posterior of the shadow model is used to train the attack model. But this requires a significant computational overhead of attack. To reduce the computational overhead, Salem \textit{et al.} \cite{salem2019ml} present a novel method to merge multiple shadow models into a single shadow model and gradually relax the adversary knowledge assumptions. 

In Addition, other researchers propose not to train the attack model but only by comparing the relationship between the size of the threshold $\tau$ set in advance and the output of the target model. Li \textit{et al.} \cite{li2021membership} present a transfer attack method by comparing the cross entropy loss of the target model output. To solve the weakness of attack based on prediction entropy, Song \textit{et al.} \cite{song2021systematic} a modified prediction entropy, combining the properties of monotonically decreasing with the correct label while monotonically increasing with the incorrect label. Salem \textit{et al.} \cite{salem2019ml} present an inference method by comparing the maximum posterior probability of the target model output. And Yeom \textit{et al.} \cite{yeom2018privacy} propose to infer the membership state by comparing the prediction loss of the samples. However, such MIAs that require only thresholds for comparison need the adversary to know the difference between the distribution of members and nonmembers in advance to design the optimal threshold for effective MIAs.

\subsection{Defense Mechanism}

The literature \cite{shokri2017membership,long2018understanding,yeom2018privacy,li20membership} point out that overfitting is one of the leading causes of the vulnerability of machine learning models to membership inference attacks. Therefore, in the research of defense against MIA, preventing machine learning models from overfitting during training is a defense below MIAs. For example, this can be achieved by $L_1$ or $L_2$ regularization \cite{li20membership}, Dropout \cite{kaya2020effectiveness}, and the robustness of the model through data augmentation \cite{kaya2021does}. Chen \textit{et al.} \cite{chen2022relaxloss} propose a gradient ascent strategy to reduce the generalization error while guaranteeing the accuracy of the target model.

In addition, there are studies to fool adversaries by perturbing the output and internal parameters of the model, thus hiding the accurate model parameters and output. Defenders can use the differential privacy stochastic gradient descent (DP-SGD) mechanism \cite{abadi2016deep}, which perturbs the data-
dependent objective function by adding noise to the gradient during the model's training, which replaces the normal gradient descent and mitigates the inference attack. Besides, Jia \textit{et al.} \cite{jia2019memguard} inspired by the adversarial sample, and propose the MemGuard mechanism, which achieves the purpose of fooling the attacker by adding perturbation noise to the output probability of the target model, so that each inference attack by the attacker is close to a random guess, thus protecting the privacy information of the training set in the target model.

\subsection{Contrastive Learning}

The success of many supervised models relies on a large amount of labeled data behind them, and obtaining labeled data is often costly or difficult to achieve. Contrastive learning is a new unsupervised learning approach that focuses on learning standard features among similar samples and distinguishing differences among non-similar samples. Instead of focusing on the tedious details of samples, contrast learning only needs to learn to distinguish data on the feature space at the abstract semantic level. Hence, the model and its optimization become more straightforward, and its generalization ability is more robust. The goal of contrastive learning is to learn an encoder that encodes similar data of the same kind and makes the encoding results of different classes of data as different as possible. Wu \textit{et al.} \cite{wu2018unsupervised} propose to achieve viewing each sample as a class by using a memory bank to store negative samples. He \textit{et al.} \cite{he2020momentum} propose using momentum contrast to do unsupervised visual representation learning. Chen \textit{et al.} \cite{chen2020simple} propose a simple visual representation contrast learning framework SimCLR, which generates negative samples in contrast learning by data augmentation techniques. All the above contrastive learning approaches require using both positive and negative samples for training models. In addition, some studies propose contrastive learning approaches that do not need negative samples. Grill \textit{et al.} \cite{grill2020bootstrap} predicted the data samples after data augmentation without using negative samples. Chen \textit{et al.} \cite{chen2021exploring} proposed a more straightforward contrastive learning framework designed by siamese networks without using negative samples.

\section{Ethic Statement}
This work aims to promote the understanding of membership inference attacks by identifying vulnerabilities in machine learning models and contributing to the development of stronger defenses. While the proposed method, CLMIA, introduces an improved attack mechanism, it is intended solely for ethical purposes, such as evaluating and enhancing model privacy and security. All experiments were conducted on publicly available datasets containing no personally identifiable information. Misuse of this work to compromise real-world systems or violate data privacy would contradict its ethical objectives. We encourage researchers and practitioners to use this work responsibly and in alignment with privacy-preserving practices.

\section{Conclusion}
\label{sec:conclusion}

In this paper, we propose an unsupervised contrastive learning approach to train an attack model for membership inference attacks. We demonstrate that training the attack model with an unsupervised contrastive learning approach has better generalization ability. Specifically, we propose a new attack method, CLMIA, which uses the output of the target model and statistically computes the features of the data identity information, and then uses the contrast learning approach to train the attack model in a black-box scenario to obtain the attack model. Our extensive experiments have confirmed that CLMIA has better attack performance, especially for the case of less marked data identity information, and CLMIA has better generalization ability. In addition, we also analyze some important factors that affect the performance of CLMIA attacks. For future work, we would like to explore more features that can distinguish data identity information and use a more lightweight contrastive learning approach to perform membership inference attacks.


\begin{thebibliography}{10}
\providecommand{\url}[1]{#1}
\csname url@samestyle\endcsname
\providecommand{\newblock}{\relax}
\providecommand{\bibinfo}[2]{#2}
\providecommand{\BIBentrySTDinterwordspacing}{\spaceskip=0pt\relax}
\providecommand{\BIBentryALTinterwordstretchfactor}{4}
\providecommand{\BIBentryALTinterwordspacing}{\spaceskip=\fontdimen2\font plus
\BIBentryALTinterwordstretchfactor\fontdimen3\font minus \fontdimen4\font\relax}
\providecommand{\BIBforeignlanguage}[2]{{%
\expandafter\ifx\csname l@#1\endcsname\relax
\typeout{** WARNING: IEEEtranS.bst: No hyphenation pattern has been}%
\typeout{** loaded for the language `#1'. Using the pattern for}%
\typeout{** the default language instead.}%
\else
\language=\csname l@#1\endcsname
\fi
#2}}
\providecommand{\BIBdecl}{\relax}
\BIBdecl

\bibitem{abadi2016deep}
M.~Abadi, A.~Chu, I.~Goodfellow, H.~B. McMahan, I.~Mironov, K.~Talwar, and L.~Zhang, ``Deep learning with differential privacy,'' in \emph{Proceedings of the 2016 ACM SIGSAC conference on computer and communications security}, 2016, pp. 308--318.

\bibitem{carlini2022membership}
N.~Carlini, S.~Chien, M.~Nasr, S.~Song, A.~Terzis, and F.~Tramer, ``Membership inference attacks from first principles,'' in \emph{2022 IEEE Symposium on Security and Privacy (SP)}.\hskip 1em plus 0.5em minus 0.4em\relax IEEE, 2022, pp. 1897--1914.

\bibitem{carlini2019secret}
N.~Carlini, C.~Liu, {\'U}.~Erlingsson, J.~Kos, and D.~Song, ``The secret sharer: Evaluating and testing unintended memorization in neural networks,'' in \emph{28th USENIX Security Symposium (USENIX Security 19)}, 2019, pp. 267--284.

\bibitem{chen2022relaxloss}
D.~Chen, N.~Yu, and M.~Fritz, ``Relaxloss: defending membership inference attacks without losing utility,'' \emph{arXiv preprint arXiv:2207.05801}, 2022.

\bibitem{chen2020practical}
H.~Chen, H.~Li, G.~Dong, M.~Hao, G.~Xu, X.~Huang, and Z.~Liu, ``Practical membership inference attack against collaborative inference in industrial iot,'' \emph{IEEE Transactions on Industrial Informatics}, vol.~18, no.~1, pp. 477--487, 2020.

\bibitem{chen2020simple}
T.~Chen, S.~Kornblith, M.~Norouzi, and G.~Hinton, ``A simple framework for contrastive learning of visual representations,'' in \emph{International conference on machine learning}.\hskip 1em plus 0.5em minus 0.4em\relax PMLR, 2020, pp. 1597--1607.

\bibitem{chen2022predicting}
X.~Chen, Y.~H. Cho, Y.~Dou, and B.~Lev, ``Predicting future earnings changes using machine learning and detailed financial data,'' \emph{Journal of Accounting Research}, vol.~60, no.~2, pp. 467--515, 2022.

\bibitem{Chen_2021_CVPR}
X.~Chen and K.~He, ``Exploring simple siamese representation learning,'' in \emph{Proceedings of the IEEE/CVF Conference on Computer Vision and Pattern Recognition (CVPR)}, June 2021, pp. 15\,750--15\,758.

\bibitem{chen2021exploring}
------, ``Exploring simple siamese representation learning,'' in \emph{Proceedings of the IEEE/CVF Conference on Computer Vision and Pattern Recognition}, 2021, pp. 15\,750--15\,758.

\bibitem{del2022leveraging}
G.~Del~Grosso, H.~Jalalzai, G.~Pichler, C.~Palamidessi, and P.~Piantanida, ``Leveraging adversarial examples to quantify membership information leakage,'' in \emph{Proceedings of the IEEE/CVF Conference on Computer Vision and Pattern Recognition}, 2022, pp. 10\,399--10\,409.

\bibitem{gao2021simcse}
T.~Gao, X.~Yao, and D.~Chen, ``Simcse: Simple contrastive learning of sentence embeddings,'' in \emph{Proceedings of the 2021 Conference on Empirical Methods in Natural Language Processing}, 2021, pp. 6894--6910.

\bibitem{grill2020bootstrap}
J.-B. Grill, F.~Strub, F.~Altch{\'e}, C.~Tallec, P.~Richemond, E.~Buchatskaya, C.~Doersch, B.~Avila~Pires, Z.~Guo, M.~Gheshlaghi~Azar \emph{et~al.}, ``Bootstrap your own latent-a new approach to self-supervised learning,'' \emph{Advances in neural information processing systems}, vol.~33, pp. 21\,271--21\,284, 2020.

\bibitem{he2020momentum}
K.~He, H.~Fan, Y.~Wu, S.~Xie, and R.~Girshick, ``Momentum contrast for unsupervised visual representation learning,'' in \emph{Proceedings of the IEEE/CVF conference on computer vision and pattern recognition}, 2020, pp. 9729--9738.

\bibitem{he2020segmentations}
Y.~He, S.~Rahimian, B.~Schiele, and M.~Fritz, ``Segmentations-leak: Membership inference attacks and defenses in semantic image segmentation,'' in \emph{European Conference on Computer Vision}.\hskip 1em plus 0.5em minus 0.4em\relax Springer, 2020, pp. 519--535.

\bibitem{hu2022membership}
H.~Hu, Z.~Sal{\v{c}}i{\'c}, G.~Dobbie, J.~Chen, L.~Sun, and X.~Zhang, ``Membership inference via backdooring,'' in \emph{Proceedings of the Thirty-First International Joint Conference on Artificial Intelligence}, vol.~23, 2022, pp. 3832--3838.

\bibitem{hui2021practical}
B.~Hui, Y.~Yang, H.~Yuan, P.~Burlina, N.~Z. Gong, and Y.~Cao, ``Practical blind membership inference attack via differential comparisons,'' in \emph{ISOC Network and Distributed System Security Symposium (NDSS)}, 2021.

\bibitem{irolla2019demystifying}
P.~Irolla and G.~Ch{\^a}tel, ``Demystifying the membership inference attack,'' in \emph{2019 12th CMI Conference on Cybersecurity and Privacy (CMI)}.\hskip 1em plus 0.5em minus 0.4em\relax IEEE, 2019, pp. 1--7.

\bibitem{jia2019memguard}
J.~Jia, A.~Salem, M.~Backes, Y.~Zhang, and N.~Z. Gong, ``Memguard: Defending against black-box membership inference attacks via adversarial examples,'' in \emph{Proceedings of the 2019 ACM SIGSAC conference on computer and communications security}, 2019, pp. 259--274.

\bibitem{kaya2021does}
Y.~Kaya and T.~Dumitras, ``When does data augmentation help with membership inference attacks?'' in \emph{International conference on machine learning}.\hskip 1em plus 0.5em minus 0.4em\relax PMLR, 2021, pp. 5345--5355.

\bibitem{kaya2020effectiveness}
Y.~Kaya, S.~Hong, and T.~Dumitras, ``On the effectiveness of regularization against membership inference attacks,'' \emph{arXiv preprint arXiv:2006.05336}, 2020.

\bibitem{leino2020stolen}
K.~Leino and M.~Fredrikson, ``Stolen memories: Leveraging model memorization for calibrated $\{$White-Box$\}$ membership inference,'' in \emph{29th USENIX security symposium (USENIX Security 20)}, 2020, pp. 1605--1622.

\bibitem{li2022user}
G.~Li, S.~Rezaei, and X.~Liu, ``User-level membership inference attack against metric embedding learning,'' \emph{arXiv preprint arXiv:2203.02077}, 2022.

\bibitem{li20membership}
J.~Li, N.~Li, and B.~Ribeiro, ``Membership inference attacks and defenses in classification models,'' in \emph{Proceedings of the Eleventh ACM Conference on Data and Application Security and Privacy}, 2021, pp. 5--16.

\bibitem{li2021membership}
Z.~Li and Y.~Zhang, ``Membership leakage in label-only exposures,'' in \emph{Proceedings of the 2021 ACM SIGSAC Conference on Computer and Communications Security}, 2021, pp. 880--895.

\bibitem{8728167}
G.~Liu, C.~Wang, K.~Peng, H.~Huang, Y.~Li, and W.~Cheng, ``Socinf: Membership inference attacks on social media health data with machine learning,'' \emph{IEEE Transactions on Computational Social Systems}, vol.~6, no.~5, pp. 907--921, 2019.

\bibitem{liu2022membership}
Y.~Liu, Z.~Zhao, M.~Backes, and Y.~Zhang, ``Membership inference attacks by exploiting loss trajectory,'' in \emph{Proceedings of the 2022 ACM SIGSAC Conference on Computer and Communications Security}, 2022, pp. 2085--2098.

\bibitem{10.1145/3548606.3560684}
\BIBentryALTinterwordspacing
------, ``Membership inference attacks by exploiting loss trajectory,'' in \emph{Proceedings of the 2022 ACM SIGSAC Conference on Computer and Communications Security}, ser. CCS '22.\hskip 1em plus 0.5em minus 0.4em\relax New York, NY, USA: Association for Computing Machinery, 2022, p. 2085–2098. [Online]. Available: \url{https://doi.org/10.1145/3548606.3560684}
\BIBentrySTDinterwordspacing

\bibitem{long2018understanding}
Y.~Long, V.~Bindschaedler, L.~Wang, D.~Bu, X.~Wang, H.~Tang, C.~A. Gunter, and K.~Chen, ``Understanding membership inferences on well-generalized learning models,'' \emph{arXiv preprint arXiv:1802.04889}, 2018.

\bibitem{miao2021audio}
Y.~Miao, M.~Xue, C.~Chen, L.~Pan, J.~Zhang, B.~Z.~H. Zhao, D.~Kaafar, and Y.~Xiang, ``The audio auditor: User-level membership inference in internet of things voice services,'' \emph{Proceedings on Privacy Enhancing Technologies}, vol.~1, pp. 209--228, 2021.

\bibitem{nasr2019comprehensive}
M.~Nasr, R.~Shokri, and A.~Houmansadr, ``Comprehensive privacy analysis of deep learning: Passive and active white-box inference attacks against centralized and federated learning,'' in \emph{2019 IEEE symposium on security and privacy (SP)}.\hskip 1em plus 0.5em minus 0.4em\relax IEEE, 2019, pp. 739--753.

\bibitem{pyrgelis2017knock}
A.~Pyrgelis, C.~Troncoso, and E.~De~Cristofaro, ``Knock knock, who's there? membership inference on aggregate location data,'' \emph{arXiv preprint arXiv:1708.06145}, 2017.

\bibitem{rahimian2021differential}
S.~Rahimian, T.~Orekondy, and M.~Fritz, ``Differential privacy defenses and sampling attacks for membership inference,'' in \emph{Proceedings of the 14th ACM Workshop on Artificial Intelligence and Security}, 2021, pp. 193--202.

\bibitem{sablayrolles2019white}
A.~Sablayrolles, M.~Douze, C.~Schmid, Y.~Ollivier, and H.~J{\'e}gou, ``White-box vs black-box: Bayes optimal strategies for membership inference,'' in \emph{International Conference on Machine Learning}.\hskip 1em plus 0.5em minus 0.4em\relax PMLR, 2019, pp. 5558--5567.

\bibitem{salem2019ml}
A.~Salem, Y.~Zhang, M.~Humbert, M.~Fritz, and M.~Backes, ``Ml-leaks: Model and data independent membership inference attacks and defenses on machine learning models,'' in \emph{Network and Distributed Systems Security Symposium 2019}.\hskip 1em plus 0.5em minus 0.4em\relax Internet Society, 2019.

\bibitem{shah2021evaluating}
M.~A. Shah, J.~Szurley, M.~Mueller, A.~Mouchtaris, and J.~Droppo, ``Evaluating the vulnerability of end-to-end automatic speech recognition models to membership inference attacks.'' in \emph{Interspeech}, 2021, pp. 891--895.

\bibitem{shokri2017membership}
R.~Shokri, M.~Stronati, C.~Song, and V.~Shmatikov, ``Membership inference attacks against machine learning models,'' in \emph{2017 IEEE symposium on security and privacy (SP)}.\hskip 1em plus 0.5em minus 0.4em\relax IEEE, 2017, pp. 3--18.

\bibitem{song2021systematic}
L.~Song and P.~Mittal, ``Systematic evaluation of privacy risks of machine learning models,'' in \emph{30th USENIX Security Symposium (USENIX Security 21)}, 2021, pp. 2615--2632.

\bibitem{song2019privacy}
L.~Song, R.~Shokri, and P.~Mittal, ``Privacy risks of securing machine learning models against adversarial examples,'' in \emph{Proceedings of the 2019 ACM SIGSAC Conference on Computer and Communications Security}, 2019, pp. 241--257.

\bibitem{wu2018unsupervised}
Z.~Wu, Y.~Xiong, S.~X. Yu, and D.~Lin, ``Unsupervised feature learning via non-parametric instance discrimination,'' in \emph{Proceedings of the IEEE conference on computer vision and pattern recognition}, 2018, pp. 3733--3742.

\bibitem{yeom2018privacy}
S.~Yeom, I.~Giacomelli, M.~Fredrikson, and S.~Jha, ``Privacy risk in machine learning: Analyzing the connection to overfitting,'' in \emph{2018 IEEE 31st computer security foundations symposium (CSF)}.\hskip 1em plus 0.5em minus 0.4em\relax IEEE, 2018, pp. 268--282.

\bibitem{zhang2022label}
G.~Zhang, B.~Liu, T.~Zhu, M.~Ding, and W.~Zhou, ``Label-only membership inference attacks and defenses in semantic segmentation models,'' \emph{IEEE Transactions on Dependable and Secure Computing}, 2022.

\bibitem{zhang2021membership}
M.~Zhang, Z.~Ren, Z.~Wang, P.~Ren, Z.~Chen, P.~Hu, and Y.~Zhang, ``Membership inference attacks against recommender systems,'' in \emph{Proceedings of the 2021 ACM SIGSAC Conference on Computer and Communications Security}, 2021, pp. 864--879.

\bibitem{zhang2022membership}
Z.~Zhang, C.~Yan, and B.~A. Malin, ``Membership inference attacks against synthetic health data,'' \emph{Journal of biomedical informatics}, vol. 125, p. 103977, 2022.

\end{thebibliography}

\onecolumn
\section*{Appendix}
\label{sec:set-diff-dodis}

\subsection{Additional Results on Different Models}
\label{A1}
\begin{table*}[htbp]
	\centering
	\caption{\textbf{Attack performance of different attacks against Simple CNN trained on three datasets.}}
	\begin{tabular}{c|c|ccccccccc}
		\toprule
		\multicolumn{2}{c}{\multirow{2}[4]{*}{\textbf{Attack method}}} & \multicolumn{3}{c}{\textbf{Balanced accuracy}} & \multicolumn{3}{c}{\textbf{$F_1$ Score}} \\
		\cmidrule(r){3-5} \cmidrule(r){6-8} \multicolumn{2}{c}{} & \multicolumn{1}{c}{CIFAR-10} & \multicolumn{1}{c}{CIFAR-100} & \multicolumn{1}{c}{STL-10} & \multicolumn{1}{c}{CIFAR-10} & \multicolumn{1}{c}{CIFAR-100} & \multicolumn{1}{c}{STL-10} \\
		
        \midrule
  	\multirow{4}[2]{*}{Based on thresholds} 
        & Top1  &0.586  &0.648 &0.611  &0.660  &0.736  &0.712   \\
		& Prediction entropy  & 0.629  &0.697  &0.671 &0.730 &0.767 &0.752    \\
		& Modified entropy & 0.626    & 0.720   &\textbf{0.716} &0.728    & 0.770   &\textbf{0.779 }   \\
		& Predicted loss  &0.581  & \textbf{0.741}&0.662    & 0.705  & \textbf{0.794}    & 0.747   \\
        \cmidrule(r){1-2}
        \multirow{1}[1]{*}{Based on labels}
        & Prediction correctness &0.613  &0.721   &0.705    &0.721   & 0.781  & 0.772    \\
        \cmidrule(r){1-2}
		\multirow{1}[1]{*}{Based on the attack model}
        & NN attack  &0.616 &0.515  &0.556 &0.731    & 0.615    &0.681    \\
         \cmidrule(r){1-2}
		\textbf{Ours} &CLMIA &\textbf{0.652} &0.738 & 0.655  &\textbf{0.757}  &0.785   &0.702  \\
		\bottomrule
	\end{tabular}%
	\label{tab:cnn_result}
\end{table*}%

\begin{table*}[htbp]
	\centering
	\caption{\textbf{Attack performance of different attacks against VGG-19 trained on three datasets.}}
	\begin{tabular}{c|c|ccccccccc}
		\toprule
		\multicolumn{2}{c}{\multirow{2}[4]{*}{\textbf{Attack method}}} & \multicolumn{3}{c}{\textbf{Balanced accuracy}} & \multicolumn{3}{c}{\textbf{$F_1$ Score}} \\
		\cmidrule(r){3-5} \cmidrule(r){6-8} \multicolumn{2}{c}{} & \multicolumn{1}{c}{CIFAR-10} & \multicolumn{1}{c}{CIFAR-100} & \multicolumn{1}{c}{STL-10} & \multicolumn{1}{c}{CIFAR-10} & \multicolumn{1}{c}{CIFAR-100} & \multicolumn{1}{c}{STL-10} \\
		
        \midrule
  	\multirow{4}[2]{*}{Based on thresholds} 
   & Top1  &0.578 &0.562 &\textbf{0.660}   &0.679   &0.695     & \textbf{0.735}    \\
	& Prediction entropy  &0.539    &0.642   & 0.563  &0.684 &0.737 &0.696   \\
	& Modified entropy & 0.545  &0.653     &0.575     &0.687      &0.742      & 0.702     \\
	& Predicted loss  &0.528  & \textbf{0.725}  & 0.554  &0.679   &0.620   & 0.695 \\
        \cmidrule(r){1-2}
        \multirow{1}[1]{*}{Based on labels}
    & Prediction correctness &\textbf{0.962}  &0.644   &0.569  &\textbf{0.963}   & 0.737  &0.698  \\
        \cmidrule(r){1-2}
		\multirow{1}[1]{*}{Based on the attack model}
  & NN attack  &0.582 &0.665  &0.619 & 0.576    & 0.772    & 0.715    \\
         \cmidrule(r){1-2}
		\textbf{Ours} &CLMIA  &0.619  & 0.679 &0.649 &0.745   &\textbf{0.784}   &0.627  \\
		\bottomrule
	\end{tabular}%
	\label{tab:vgg19_result}
\end{table*}%


\end{document}